\documentclass[conference]{IEEEtran}
\IEEEoverridecommandlockouts
\usepackage{cite}
\usepackage{amsmath,amssymb,amsfonts}
\usepackage{algorithmic}
\usepackage{graphicx}
\usepackage{textcomp}
\usepackage{xcolor}
\usepackage{hyperref}
\def\BibTeX{{\rm B\kern-.05em{\sc i\kern-.025em b}\kern-.08em
    T\kern-.1667em\lower.7ex\hbox{E}\kern-.125emX}}
\begin{document}

\title{RGAT: A Deeper Look into Syntactic Dependency Information for Coreference Resolution\\
}

\author{\IEEEauthorblockN{Yuan Meng, Xuhao Pan, Jun Chang$^{\ast}$ \thanks{*Corresponding author}, Yue Wang}
\IEEEauthorblockA{\textit{School of Statistics and Information} \\
\textit{Shanghai University of International Business and Economics}\\
Shanghai, China \\
\{nancymeng, 21332006, changjun\}@suibe.edu.cn, wy416408@foxmail.com}\\
}

\maketitle

\begin{abstract}
Although syntactic information is beneficial for many NLP tasks, combining it with contextual information between words to solve the coreference resolution problem needs to be further explored. In this paper, we propose an end-to-end parser that combines pre-trained BERT \cite{b1} with a Syntactic Relation Graph Attention Network (RGAT) to take a deeper look into the role of syntactic dependency information for the coreference resolution task. In particular, the RGAT model is first proposed, then used to understand the syntactic dependency graph and learn better task-specific syntactic embeddings. An integrated architecture incorporating BERT embeddings and syntactic embeddings is constructed to generate blending representations for the downstream task. Our experiments on a public Gendered Ambiguous Pronouns (GAP) dataset show that with the supervision learning of the syntactic dependency graph and without fine-tuning the entire BERT, we increased the F1-score of the previous best model (RGCN-with-BERT) \cite{b2} from 80.3\% to 82.5\%, compared to the F1-score by single BERT embeddings from 78.5\% to 82.5\%. Experimental results on another public dataset – OntoNotes 5.0 demonstrate that the performance of the model is also improved by incorporating syntactic dependency information learned from RGAT.
\end{abstract}


\section{Introduction}
Coreference resolution is the task of finding all linguistic expressions that refer to the same entity in the natural language. Ambiguous pronoun resolution, which attempts to resolve gendered ambiguous pronouns in English such as ‘he’ and ‘she’, is a longstanding challenge in coreference resolution \cite{b2}. A Kaggle competition based on the task of gendered ambiguous pronouns (GAP) resolution was conducted in 2019 \cite{b3}. The effective use of Bidirectional Encoder Representations from Transformers or BERT \cite{b1} in this competition has shown significant improvement over traditional approaches. Unlike the traditional unidirectional language model, BERT is designed to pre-train deep bidirectional representations using a new masked language model (MLM), which enables the generation of deep bidirectional contextual embeddings.

At present, there are two BERT-based approaches for applying these contextual embeddings to ambiguous pronoun resolution tasks: the feature-based approach using BERT and fine-tuning BERT approach. The feature-based approach using BERT treats contextual representations derived from BERT as extra input features, which are combined in a task-specific model architecture without fine-tuning any parameters of BERT to obtain the coreference resolution for the target pronoun. For example, a model architecture combining BERT and SVM proposed in \cite{b4} obtains the correct mention for the target pronoun by applying the contextual embeddings from BERT to an SVM classifier. As for fine-tuning BERT approach, it uses BERT to model the downstream gendered pronoun reference task by plugging in the task-specific inputs and outputs into BERT and fine-tuning all the parameters end-to-end. Compared to the feature-based approach using BERT, fine-tuning BERT approach obtains more impressive performance without considering the computational cost, such as single fine-tuned BERT \cite{b5} or ensemble learning from multiple fine-tuned base BERT models \cite{b6}. 

However, fine-tuning the entire BERT model for a specific task is very computationally expensive and time-consuming because all parameters are jointly fine-tuned on the downstream task and need to be saved in a separate copy. For this reason, there are two improving strategies in BERT-based approach for the gendered pronoun reference task. One strategy focuses on the output representation of each layer in BERT by altering the BERT structure slightly at each layer and adding some extra parameters to change the output of each layer. Compared to fine-tuning all the parameters of BERT, this strategy can obtain a better result with less computation time, like Adapter \cite{b7} , LoRA \cite{b8} and so on. Another strategy is to explore better blending embeddings than BERT on the coreference task with the help of syntactic parsing information. Syntactic parsing information is a strong tool in many NLP tasks, such as entity extraction or relation extraction. It has also been verified that blending embeddings from BERT representations and syntactic embeddings outperform the original BERT contextual representations in the gendered pronoun reference task \cite{b2}. Since the strategy of exploring blending embeddings has a computational advantage in running many experiments with cheaper models on a pre-compute representation of BERT, it is worthwhile for us to explore again the value of blending embeddings incorporating syntactic dependency information in ambiguous pronoun resolution tasks. 

Recently, Cen et al. \cite{b9} have proposed the GATNE model, which is a large-scale heterogeneous graph representations learning model to effectively aggregate neighbors of different edge types to the current node by assigning an attention mechanism. As far as we know, there has been no study has attempted to use GATNE or its variants to digest heterogeneous graph structures from the syntactic dependency graph.

Inspired by the GATNE model, we propose our Syntactic Relation Graph Attention Network model to make it suitable to generate heterogeneous syntactic embeddings for each sample data, namely RGAT. Based on that, we propose an end-to-end solution by combining pre-trained BERT with RGAT. Experiment results on the public GAP (Gendered Ambiguous Pronouns) dataset released by Google AI demonstrate that the blending embeddings which combine BERT representations and syntactic dependency graph representations outperform the original BERT-only embeddings on the pronoun resolution task, which significantly improves the baseline F1-score from 78.5\% to 82.5\% without fine-tuning BERT and expensive computing resource. Furtherly, to verify the effectiveness of our RGAT model for digesting syntactic dependency information in coreference resolution tasks, we also conduct another coreference resolution experiment on the public NLP dataset--OntoNotes 5.0 dataset. The experiment results demonstrate that after the syntactic embeddings learned with our RGAT model are incorporated with the benchmark model, the F1-score improves from 76.9\% to 77.7\%. All our experiment codes in this paper are available at \href{https://github.com/qingtian5/RGAT_with_BERT}{https://github.com/qingtian5/RGAT\_with\_BERT}. Our main contributions are shown below:
\begin{itemize}
    \item Our work is the first deep attempt at using heterogeneous graph representations learning with attention mechanism on syntactic dependency graph for pronoun resolution task. The syntactic embeddings derived from our RGAT model successfully boost the performance of BERT-only embeddings. This provides a new idea to further digest syntactic dependency information for reference resolution tasks.
    \item Our work is the first to use graph attention mechanism to learn small syntactic dependency graph embeddings without expensive computation cost to solve the coreference resolution task. The supplementary experiment result on the public GAP dataset and OntoNotes 5.0 dataset shows that our adjusted model RGAT has a better generalization ability in NLP coreference resolution tasks.
    \item Our work is the first to largely boost the performance of the ambiguous pronoun resolution task with the help of syntactic dependency information. Most previous research considers that the effect of syntactic embeddings is weak, but our work significantly improves the F1-score of the BERT + fc baseline model from 78.5\% to 82.5\% on the GAP dataset.
\end{itemize}


\section{Preliminary Wrok}
\subsection{BERT-Based Embeddings}
BERT makes use of Transformer, an attention mechanism that learns contextual relations between words in a text. When training language models, BERT uses two training strategies:  Masked LM (MLM) and Next Sentence Prediction (NSP). Through these two tasks, what we need to do is how to apply BERT model to our samples to get the embedded representations.

\begin{figure}[htbp]
\centerline{\includegraphics[width=1.0\linewidth]{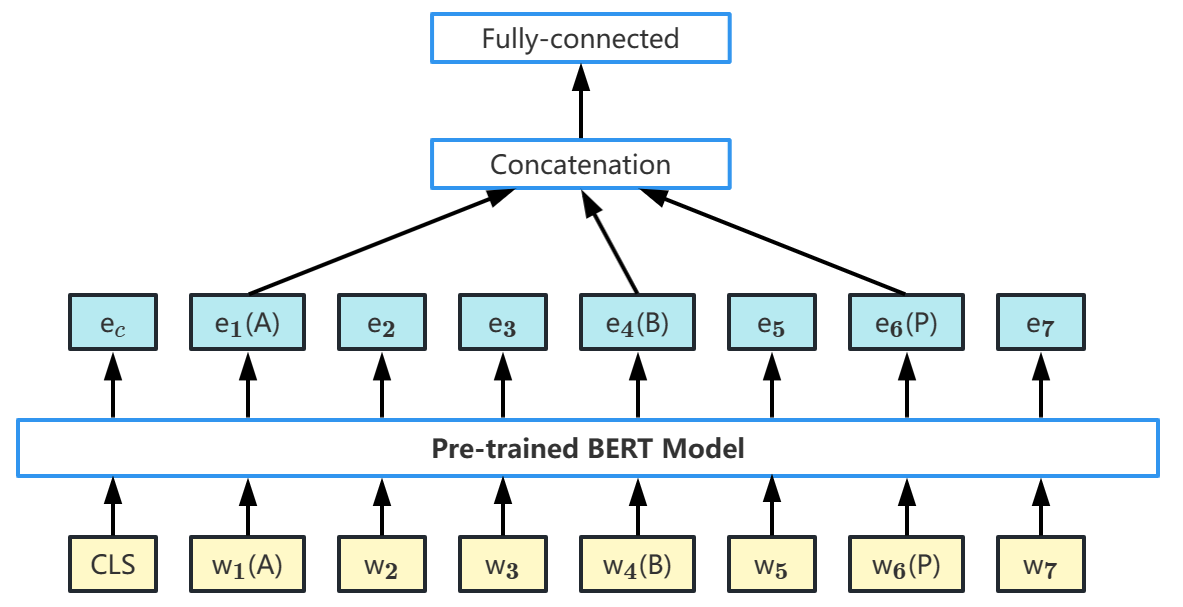}}
\caption{BERT-Based Embedding for our coreference resolution task.}
\label{fig1}
\end{figure}

For our ambiguous pronoun resolution task, each of our samples is taken as a long sentence, and then a [cls] token is added before the sentence. Through a pre-trained BERT model, the embedded representation of each token in the sentence is obtained. In fact, our goal is to obtain the relations between pronouns and nouns, so we only need to extract the embedded representations of the tokens related to pronouns (P) and nouns (A, B), then concatenate them, and finally get the results about the specific reference of the pronouns (P) through the fully connected layer, which are shown in Fig.~\ref{fig1}.

Take a sample sentence for example-- “Bill(A) said Alice(B) would arrive soon, and she(P) did”, our task is to find out whether “she(P)” refers to “Bill(A)” or “Alice(B)”. As the information flows in Fig. 1, we first break the whole sentence into words and make them the input of the BERT model. The pre-trained BERT model will generate an embedding for each word. Because in our coreference resolution task, there are only three possible results: (1) P refers to A; (2) P refers to B; (3) P refers to neither A nor B. Therefore, we regard our task as a tripartite classification problem. Thus, we extract the embedding of P, A and B from the BERT outputs, then concatenate them, and last through a fully connected layer. 

\subsection{Syntactic Dependency Information Learning}
Although syntactic parsing information is beneficial to pronoun coreference resolution, how to extract syntactic embeddings and incorporate them with BERT embeddings for the coreference task is difficult. A common way of digesting syntactic parsing information is to utilize the syntactic dependency relations between words in a text, which can be easily represented as nodes and edges in a graph structure. Then graph based model can be used to learn the syntactic dependency information for the subsequent task. For the coreference resolution task, each sentence is parsed into a syntactic dependency graph, which contains three types of edges. Thus, the traditional Graph Convolutional Network (GCN) cannot handle this multi-relation graph. Xu \cite{b2} innovatively incorporated syntactic embeddings, which is digested with Gated Relational Graph Convolutional Network (Gated RGCN \cite{b10}) with BERT embeddings for their pronoun coreference task. Specifically, RGCN is used to aggregate three heterogeneous graph structures between the head world and the dependency word to obtain word syntactic embeddings \cite{b2}. The idea provided by RGCN is that the information should be treated differently for different edge types, denoted as follows:

\begin{equation}
h_{i}^{(l+1)}=\operatorname{ReLU}\left(\sum_{r \in R} \sum_{u \in N_{r}\left(v_{i}\right)} \frac{1}{c_{i, r}} W_{r}^{(l)} h_{u}^{(l)}\right)
\end{equation}
where $N_{r}\left(v_{i}\right)$ and $W_{r}^{(l)}$ denote the set of neighbors of node i and weight under relation $r \in R$ respectively. It can be seen from here that although RGCN is used to solve multilateral types, it does not consider the problem of edge features, and the default is that only type feature exists for each edge.

In contrast to using pre-trained BERT embeddings and fully-connected layers for prediction, the series connection architecture of pre-trained BERT with RGCN from Xu \cite{b2} increases the F1-score by 1.8\%. However, RGCN does not perform very well in digesting the weight information between multiple edges graph structures from the syntactic dependency graph. Meanwhile, Xu’s result is far less accurate than fine-tuning the entire BERT. The main problem may exist in two aspects. On the one hand, according to the RGCN model, if there are multiple different types of edges in the network, it will eventually need to generate a linear layer for each type of edge. This will lead to a linear increase in the number of model parameters. On the other hand, for some types of edge in syntactic dependency graph, the frequency of occurrence may be small, which will lead to the linear layer corresponding to this type of edge is eventually updated on only a few nodes, resulting in the problem of overfitting a small number of nodes.

Inspired by RGCN with BERT and the development of graph neural networks, we believe that the performance of syntactic parsing information on pronoun resolution can be further improved. The first reason is that syntactic information always plays a very important role in the extraction of hand-crafted features, especially in most heuristics-based methods for the pronoun resolution task \cite{b11} \cite{b12} \cite{b13}. The second reason is that many newly graph learning-based models incorporating syntactic information achieve improving results in entity extraction \cite{b14} or semantic role labelling tasks \cite{b15}. In order to solve the problems of RGCN, we will illustrate how to learn the syntactic dependency graph using our proposed RGAT model in the next section. And in section IV, we propose to use L2 regularization for parameters in the RGAT model to alleviate the problem of overfitting.

\section{Method}
\subsection{Syntactic Dependency Graph}
\begin{figure}[htbp]
\centerline{\includegraphics[width=1.0\linewidth]{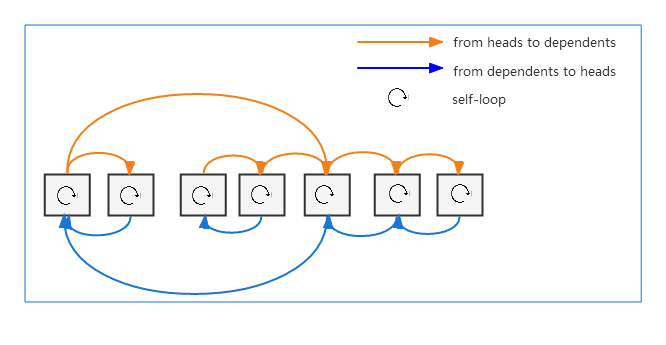}}
\caption{Information Flows in Syntactic Dependency Graph.}
\label{fig2}
\end{figure}

Since a dependency parse describes the syntactic relations that hold among words, many existing researches transform the dependency parse tree into the syntactic dependency graph to capture the syntactic features \cite{b2}. It is commonly assumed that there are three kinds of information flows in the syntactic dependency graph: from heads to dependents, from dependents to heads and self-loops, which are shown in Fig.~\ref{fig2}.

For each node in the syntactic dependency graph in Fig.~\ref{fig2}, it is linked with three different types of edges, corresponding to three different types of syntactic relations. Since we are focused on embedding learning in the syntactic dependency graph, it is important to be able to draw on strengths from such different relations and learn uniform syntactic embeddings.

\subsection{RGAT Model}
The core idea of GATNE-T (we refer to this as GATNE in the paper) is to aggregate neighbors of different edge types to the current node and then generate different vector representations for nodes of each edge type. Inspired by the GATNE algorithm proposed by Cen \cite{b9}, we adjust GATNE and propose the RGAT model applied in the syntactic dependency graph with multiple edges to learn uniform embeddings.

Generally, the RGAT model has been modified in three aspects based on the GATNE model. First, GATNE obtains the embedded representations of nodes and edges on a large graph structure data, but our RGAT model needs to adapt to different syntactic graph structures generated by different samples. Therefore, a new attention architecture is proposed to solve this problem. Second, the initial embedding representations of GATNE are randomly generated, but our goal is to solve the ambiguous pronouns coreference resolution, thus it is natural to use BERT Embeddings to initialize the node embeddings. Third, in order to take advantage of all the information, we add a shortcut module in the model so that our initialization node embeddings can also be concatenated into the final output embeddings.

Specially, the RGAT model splits the overall embedding of a certain node v on each edge type i in the syntactic dependency graph into two parts: base embedding and three edge embeddings. The base embedding of node v  is shared between its three different edge types. We use BERT Embedding as the base embedding of the nodes. We follow the following aggregation steps to obtain the final syntactic embeddings of each node.

First, the representation of each node is compressed to obtain a more compact representation denoted as $u_{i}^{\text {base }}$, which is used as the base embedding.
\begin{equation}
u_{i}^{\text {base }}=W_{r 0} u_{i}^{\text {out }}
\end{equation}
where $W_{r 0} \in R^{1024 * 256}$ is learnable, and $u_{i}^{\text {out }}$ is the BERT representation of node $v$ on edge type $i$.  In our work, the representation of each node $v$ from BERT is a vector of 1024 dimensions. Consistent with previous work \cite{b2}, the compressed node representation dimensions are set to 256. So $u_{i}^{\text {base }}$ is a vector of 256 dimensions.

Second, following GraphSage \cite{b16}, we obtain each type of edge embedding for node $v$ by aggregating from neighbors’ edge embeddings. We randomly sample $n$ neighbor nodes for each edge embedding $u_{j,r}^{\text {base }}$, and then aggregate them. The aggregator function can be a sum, mean or max-pooling aggregator as follows:
\begin{equation}
U_{i, r}=W_{r 1} \text {aggregator}\left(\left\{u_{j, r}^{\text {base}}, \forall u_{j} \in N_{i, r}, j=0,1,2, \ldots, n\right\}\right)
\end{equation}
where $W_{r 1} \in R^{d*m}$, is a learnable parameter, $d$ is 256, $m$ is the hyperparameter that needs to be given. In order to make the attention calculation more convenient, we compress the aggregated representation again.

Third, applying the attention mechanism to get the weight of each aggregated edge representation for each node as follows:
\begin{equation}
    a_{i, r}=\operatorname{softmax}\left(w_{r}^{T} \tanh \left(W_{r 2} U_{i, r}\right)\right)^{T}
\end{equation}
where $W_{r} \in R^{n}$, $W_{r 2} \in R^{m*n}$ are learnable parameters, $n$ is the hyperparameter that needs to be given.

Fourth, combining each weighted aggregated representation with the base embedding, the final representation of each node in edge type $r$ can be represented as $v_{i,r}$, which is denoted as follows:
\begin{equation}
    v_{i, r}=u_{i}^{\text {base}}+a_{i, r} M_{r} U_{i, r}
\end{equation}
where $M_{r} \in R^{d*m}$ is a learnable parameter, $u_{i}^{\text {base }}$ is base embedding and $v_{i, r}$  is a vector with 256 dimensions.

Finally, the syntactic embedding representation of each node is aggregated from three kinds of node representation $v_{i,0}$,$v_{i,1}$,$v_{i,2}$, denoted as follows:
\begin{equation}
    v_{i}=\text {aggregator}\left(v_{i, 0}, v_{i, 1}, v_{i, 2}\right)
\end{equation}
where the aggregator function can be a sum, mean or concatenate operation. The influence of different aggregators will be mentioned in detail in the ablation experiment in Section IV.

\subsection{Syntactic Embeddings}
The previous RGCN model \cite{b2} only uses the information of neighbor nodes, but it brings a significant improvement in coreference resolution tasks. Therefore, we think the potential of learning syntactic structure using RGAT can be much more than that. The most important reason is that by designing attention mechanisms, we obtain more valuable information from different types of edges. 

\begin{figure}[htbp]
\centerline{\includegraphics[width=1.0\linewidth]{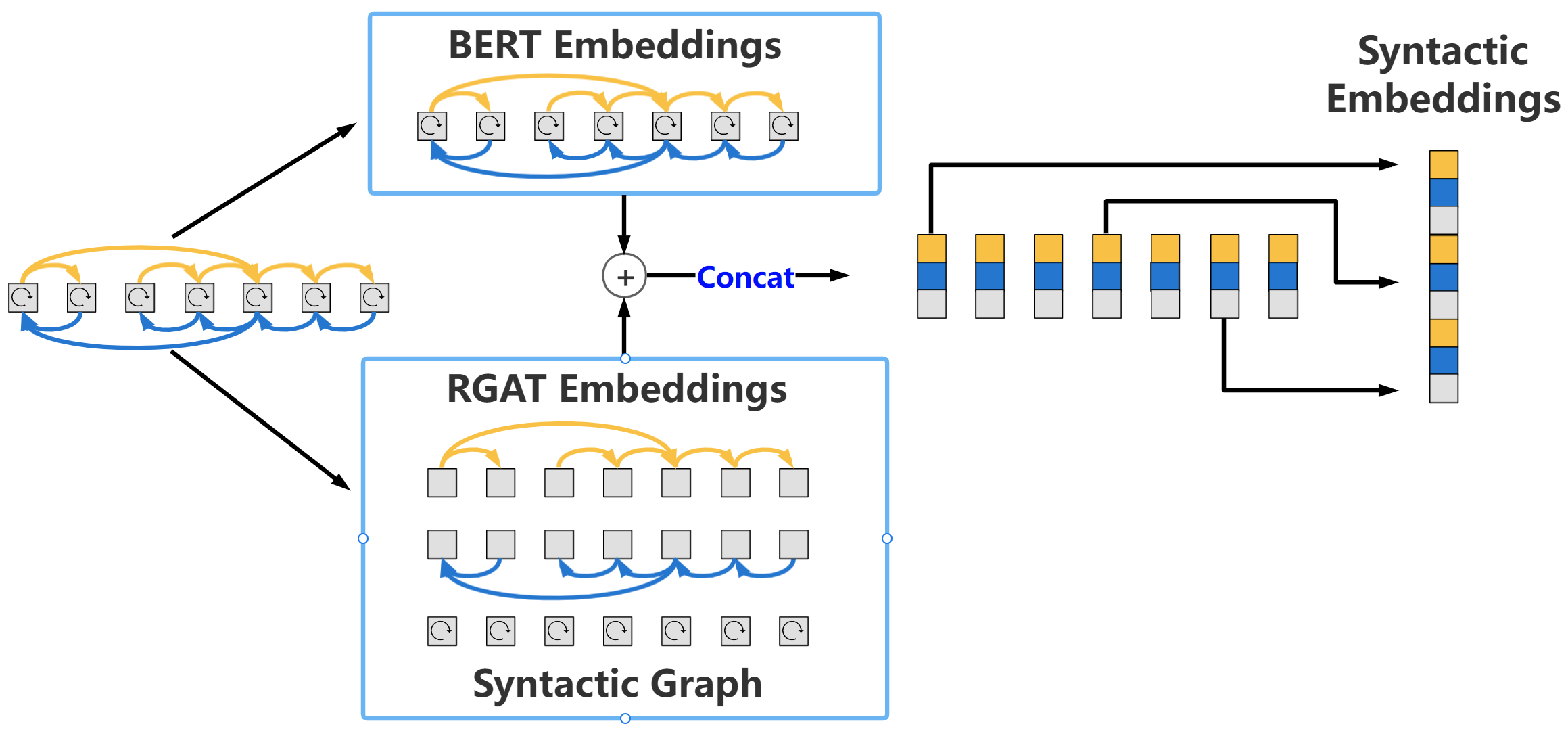}}
\caption{Embedding Structure of Syntactic Dependency Graph.}
\label{fig3}
\end{figure}

Fig.~\ref{fig3} explains in detail how to extract Syntactic Embeddings from information flows in syntactic dependency graph and BERT Embeddings. 

For one thing, we use BERT Embeddings as the base embeddings. For another thing, we use three different types of syntactic relations to construct the syntactic relation graph with attention information to learn RGAT embeddings. In order to retain information from BERT Embeddings, we concatenate the embeddings that represent the relation graph with attention information and BERT Embeddings. Then, we concatenate different embeddings from three different kinds of edges (different color represent embeddings from different edge types in Fig.~\ref{fig3}). Finally, the significant words embeddings (the embedding of three words – A, B, P, like in section II, Fig.~\ref{fig1}) are concatenated as Syntactic Embedding.

\subsection{Connect BERT Embeddings and Syntactic Embeddings in Series}
We blend the syntactic embedding derived from the syntactic dependency graph with the pretrained BERT embeddings by connecting BERT embedding and syntactic embedding in series. This integrated architecture can help us learn better-performing embeddings when dealing with the task of pronoun resolution.

\begin{figure}[htbp]
\centerline{\includegraphics[width=1.0\linewidth]{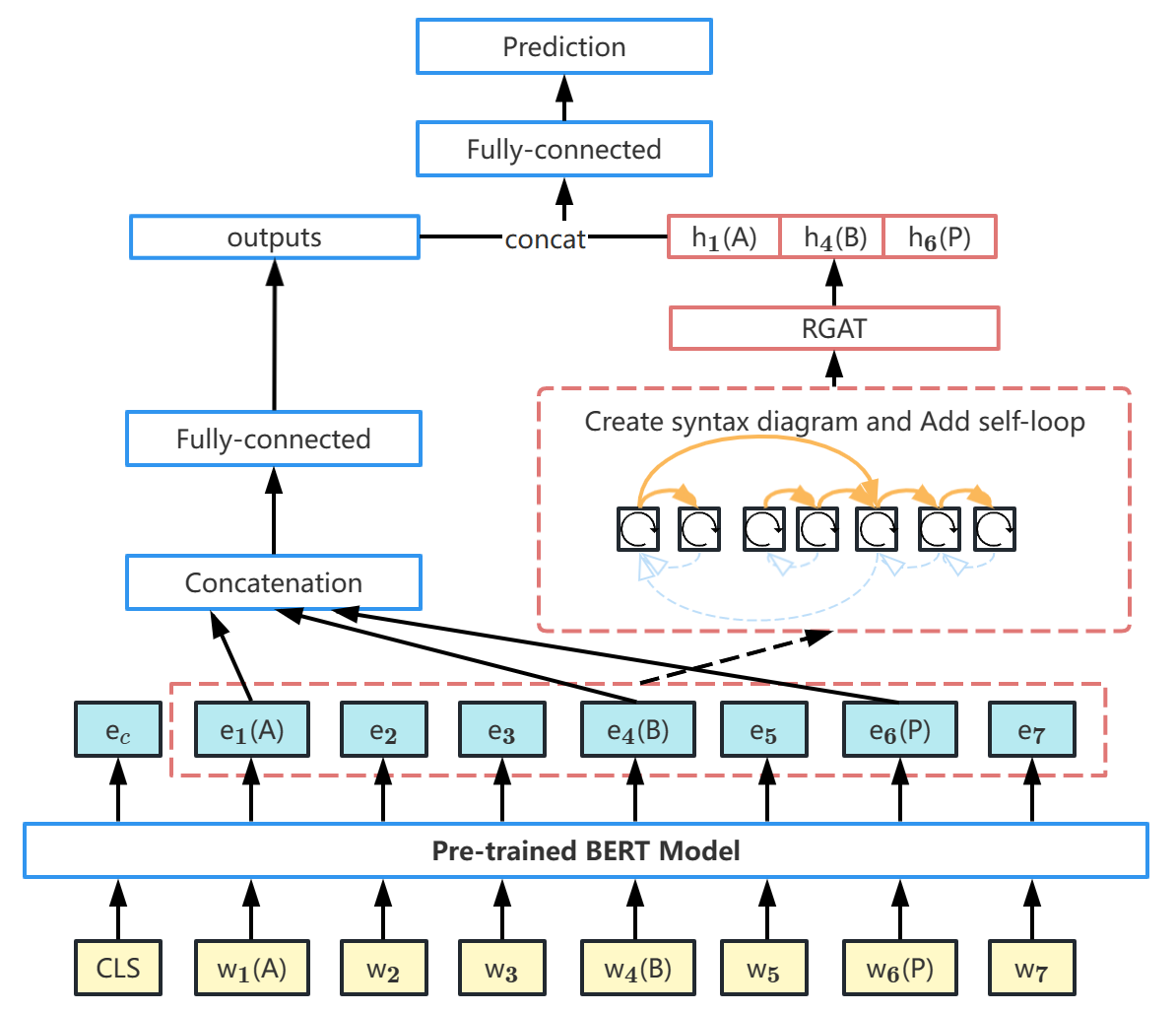}}
\caption{The Blending Structure of BERT and Syntactic Embedding.}
\label{fig4}
\end{figure}

We first use the pre-trained BERT to extract context information between words and then connect with syntactic information from RGAT to form a “look again” mechanism to further obtain blending representations that are more beneficial to the current task. The specific architecture is shown in Fig.~\ref{fig4}.

As shown in Fig.~\ref{fig4}, the pre-trained BERT obtains the hidden feature representation, then RGAT looks at the syntactic information of the sentence again. Relying on the syntactic information derived from RGAT, we can obtain the hidden state of pronoun-related words (denoted as h1(A), h4(B), h6(P)) in the sentence. There is also a fully connected layer in parallel with the output of RGAT, which is used to get a more compact embedding representation for each pronoun-related word.

Finally, the outputs representation by RGAT are concatenated with the compact embedding representation of each pronoun-related word. The reason for concatenation is mainly because the syntactic dependency graph uses a special form of Laplace smoothing during its construction process \cite{b17}, which may contain vertex-related features. Some original feature embeddings can be preserved by concatenation, and ultimately a fully connected layer is used for prediction.

\section{Experiments}
\subsection{Experimental Setup}
\noindent \textbf{GAP Dataset.} The first ACL workshop on Gender Bias in Natural Language Processing (2019) included a coreference task addressing Gendered Ambiguous Pronouns (GAP). The task is based on the coreference challenge defined by Webster \cite{b3} \cite{b18} and aims to benchmark the ability to address pronouns in real-word contexts in a gender-equitable way. 263 teams competed through the Kaggle competition, and the winning system ran at a speed of 0.13667, which is close to gender parity. We reviewed the approaches and their documentation of the top eleven systems, noting that they effectively use BERT [1], both through fine-tuning, feature extraction, or ensembles. In order to compare with the baseline results of the previous work on the GAP task \cite{b2} \cite{b3} \cite{b18} our work directly uses the Gendered Ambiguous Pronouns (GAP) dataset, containing all 8908 samples. The dataset contains 8908 pairs of labelled samples from Wikipedia. Consistent with previous work, 4908 samples are used as training data and 4000 samples are used as test data.

\noindent \textbf{Evaluation metrics.} The task is a multi-classification problem, and we use micro F1-score as the evaluation metric, which is to calculate the precision and recall of all classes together, and then calculate the micro F1-score according to the following formula. 

\begin{equation}
    F 1=2 \times \frac{\text { precision } \times \text { recall }}{\text { precision }+ \text { recall }}
\end{equation}
\begin{equation}
    \text {precision}_{\text {micro}}=\frac{T P_{1}+T P_{2}+T P_{3}}{T P_{1}+F P_{1}+T P_{2}+F P_{2}+T P_{3}+F P_{3}}
\end{equation}
\begin{equation}
    \text {recall}_{\text {micro}}=\frac{T P_{1}+T P_{2}+T P_{3}}{T P_{1}+F N_{1}+T P_{2}+F N_{2}+T P_{3}+F N_{3}}
\end{equation}
\begin{equation}
    \text {micro } F 1-\text {score}=2 \times \frac{\text {precision}_{\text {micro}} \times \text {recall}_{\text {micro}}}{\text {precision}_{\text {micro}}+\text {recall}_{\text {micro}}}
\end{equation}

\noindent \textbf{Implementation Detail.} We use the SpaCy module as a syntactic dependency analyzer in our work. For each sample, we build a syntactic dependency graph, then extract and save the needed information. Due to memory constraints, we do not put the entire syntactic dependency graph into the model for training, but first extract the features we needed from the graph, then combine them into a batch, and last sent them into the model. We use Adam \cite{b19} as the optimizer and adopt the form of warm up for the learning rate in our model. Especially, the L2 regularization result of the RGAT layer weight is added to the loss function, and the fully connected layer uses batch normalization and dropout strategy. In addition, for the number of sampling random neighbor nodes in the second step of the RGAT model, we found that each node has at most four different neighbors. Thus, in order to take into account the algorithm efficiency and the diversity of neighbor node calculations, we set the number of random samples to four. As for the aggregation method involved in the second step of GraphSage \cite{b16}, we found that methods such as summation, mean, and maximum pooling basically have no different impact on the model performance. So, we simply use the aggregation method of summation. We use the “BERT-Large-Uncased” model to generate the BERT embedding representations we need. It should be noted that in our model, BERT has not been fine-tuned. The parameters of all BERT models are fixed during training. The advantage of this is that we do not need to save a separate copy of BERT-related model parameters for the GAP dataset.

In order to better improve the generalization of the model, we used the 5-fold cross validation method to split the training set into five equal parts.  Each experiment takes one part for verification and the rest is used for training. Each time, the model parameters with the best performance on the validation set are applied to the test set to get the prediction result. A total of five prediction results are obtained, and the average value is taken as the final prediction result.

\subsection{Ablation Studies}

\begin{table}[htbp]
\caption{Ablation Experiment on GAP Dataset}
\begin{center}
\begin{tabular}{|c|c|c|c|}
\hline
\textbf{Model} & \textbf{Type dim (m, n)}& \textbf{The way of link}& \textbf{F1-score} \\
\hline
\textbf{RGAT-with-} & 5,10 & Mean/Sum & 81.5\% \\
\cline{2-4}
\textbf{BERT} & 10,20 & Mean/Sum & 81.7\% \\
\cline{2-4}
\textbf{(ours)} & 30,60 & Mean/Sum & 80.9\% \\
\cline{2-4}
 & 5,10 & Concat & 82.3\% \\
\cline{2-4}
 & 10,20 & Concat & \textbf{82.5\%} \\
\cline{2-4}
 & 30,60 & Concat & 81.1\% \\
\hline
\end{tabular}
\label{tab1}
\end{center}
\end{table}

In order to be consistent with baseline work, we use the same hyperparameter configuration as RGCN-with-BERT \cite{b2}. However, in our proposed RGAT model, it is necessary to determine the dimension of the embeddings of the edge. To this end, we conduct experiments on different parameters and the results are shown in the TABLE~\ref{tab1}.

Through comparison, the dimension parameters (m, n) of the node type are set to (10.20). For the output features of three different edge types, we compare the F1-score of different aggregation methods such as averaging, summing and concatenation. Meanwhile, since the concatenation method will lead to an increase in the number of parameters, we adjust the feature dimension so that the parameters of three aggregation methods are relatively close. In the end, the mean and summing aggregation methods are found to obtain similar experiment results, while the concatenation aggregation method is found to obtain the best result.

\subsection{Comparison with Other Methods}
The paper proposing the GAP dataset \cite{b3} \cite{b18} introduced several baseline methods: (1) existing parsers including a rule-based system by Lee \cite{b20}, as well as three neural parsers from Clark and Manning (2015) \cite{b21}, Wiseman et al. (2016) \cite{b22} and Lee et al. (2017) \cite{b23}. (2) baselines based on traditional coreference cues; (3) baselines based on structural cues: syntactic distance and parallelism; (4) baselines based on Wikipedia cues; (5) Transformer models \cite{b24}. Among them, RGCN-with-BERT \cite{b2} further improves the F1-score of some baseline methods, reaching 80.3\%.

We select the best three from the baseline models and the RGCN-with-BERT (Xu et al., 2019) \cite{b2} model to compare with our model. At the same time, we also compare our work with the BERT in series with a fully connected layer (BERT-fc). Experimental results show that our work achieves a large improvement over baseline models, which is shown in TABLE~\ref{tab2}.

\begin{table}[htbp]
\caption{Results on GAP Dataset}
\begin{center}
\begin{tabular}{|c|c|c|}
\hline
 & \textbf{Model} & \textbf{F1-score} \\
\hline
\textbf{baseline} & Lee et al. (2017) \cite{b23} & 64.0\% \\
\cline{2-3}
 & Parallelism \cite{b23} & 66.9\% \\
\cline{2-3}
 & Parallelism + URL \cite{b23} & 70.6\% \\
\hline
\textbf{BERT-baseline} & Bert + fc \cite{b2} & 78.5\% \\
\cline{2-3}
 & RGCN-with-BERT \cite{b2} & 80.3\% \\
\hline
\textbf{ours} & RGAT-with-BERT & \textbf{82.5\%} \\
\hline
\end{tabular}
\label{tab2}
\end{center}
\end{table}

Specially, our RGAT model further leverages the structure of syntactic graphs and feature extraction information for specific referential tasks. Without fine-tuning the parameters of the original BERT model, the F1-score of the task is greatly improved from 80.3\% of the previous best model to 82.5\%. Compared with BERT + fc (details are in section III), it has increased from 78.5\% to 82.5\%.

\subsection{RGAT Model Verification on OntoNotes 5.0 Dataset}
\noindent \textbf{OntoNotes dataset.} OntoNotes 5.0 (English) is a document-level dataset from the CoNLL-2012 shared task on coreference resolution. It consists of about one million words of newswire, magazine articles, broadcast news, broadcast conversations, web data and conversational speech data, and the New Testament.

\noindent \textbf{Evaluation metrics.} The main evaluation is the average F1 of three metrics -- $MUC$, $B^3$ and $CEAF_{\varphi 4}$ on the test set according to the official CoNLL-2012 evaluation scripts. We briefly describe how to calculate these three metrics.

\noindent For $MUC$, we can calculate its recall and precision as follows:
\begin{equation}
    \operatorname{recall}(R)=\frac{\sum_{i=1}^{N_{k}}\left(\left|K_{i}\right|-\left|p\left(K_{i}\right)\right|\right)}{\sum_{i=1}^{N_{k}}\left(\left|K_{i}\right|-1\right)}
\end{equation}
\begin{equation}
    \operatorname{precision}(P)=\frac{\sum_{i=1}^{N_{r}}\left(\left|R_{i}\right|-\left|p^{\prime}\left(R_{i}\right)\right|\right)}{\sum_{i=1}^{N_{r}}\left(\left|R_{i}\right|-1\right)}
\end{equation}
where $K_i$ is a collection of pronouns referring to the noun $i$, $R_i$ is the set of pronouns that the model outputs referring to noun $i$, $p(K_i)$ and $p^{\prime}(R_i)$ represents the intersection of $K_i$ and $R_i$. F1-score can be calculated by formula (7).

\noindent For $B^3$, we can calculate its recall and precision as follows:
\begin{equation}
    \operatorname{recall}(R)=\frac{\sum_{i=1}^{N_{k}} \sum_{j=1}^{N_{r}} \frac{\left|K_{i} \cap R_{j}\right|^{2}}{\left|K_{i}\right|}}{\sum_{i=1}^{N_{k}}\left|K_{i}\right|}
\end{equation}
\begin{equation}
    \operatorname{precision}(P)=\frac{\sum_{i=1}^{N_{k}} \sum_{j=1}^{N_{r}} \frac{\left|K_{i} \cap R_{j}\right|^{2}}{\left|K_{i}\right|}}{\sum_{j=1}^{N_{r}}\left|R_{j}\right|}
\end{equation}

\noindent For $CEAF_{\varphi 4}$, we can calculate its recall and precision as follows:
\begin{equation}
    \operatorname{recall}(R)=\frac{\sum_{i=1}^{N_{k}} \frac{2\left|K_{i} \cap R_{i}\right|}{\left|K_{i}\right|+\left|R_{i}\right|}}{\sum_{i=1}^{N_{k}}\left|K_{i}\right|}
\end{equation}
\begin{equation}
    \operatorname{precision}(P)=\frac{\sum_{i=1}^{N_{r}} \frac{2\left|K_{i} \cap R_{i}\right|}{\left|K_{i}\right|+\left|R_{i}\right|}}{\sum_{i=1}^{N_{r}}\left|R_{i}\right|}
\end{equation}

\noindent where $K_i$ is a collection of pronouns referring to the noun $i$, $R_i$ is the set of pronouns that the model outputs referring to noun $i$. F1-score can be calculated by formula (7).

\begin{figure}[htbp]
\centerline{\includegraphics[width=1.0\linewidth]{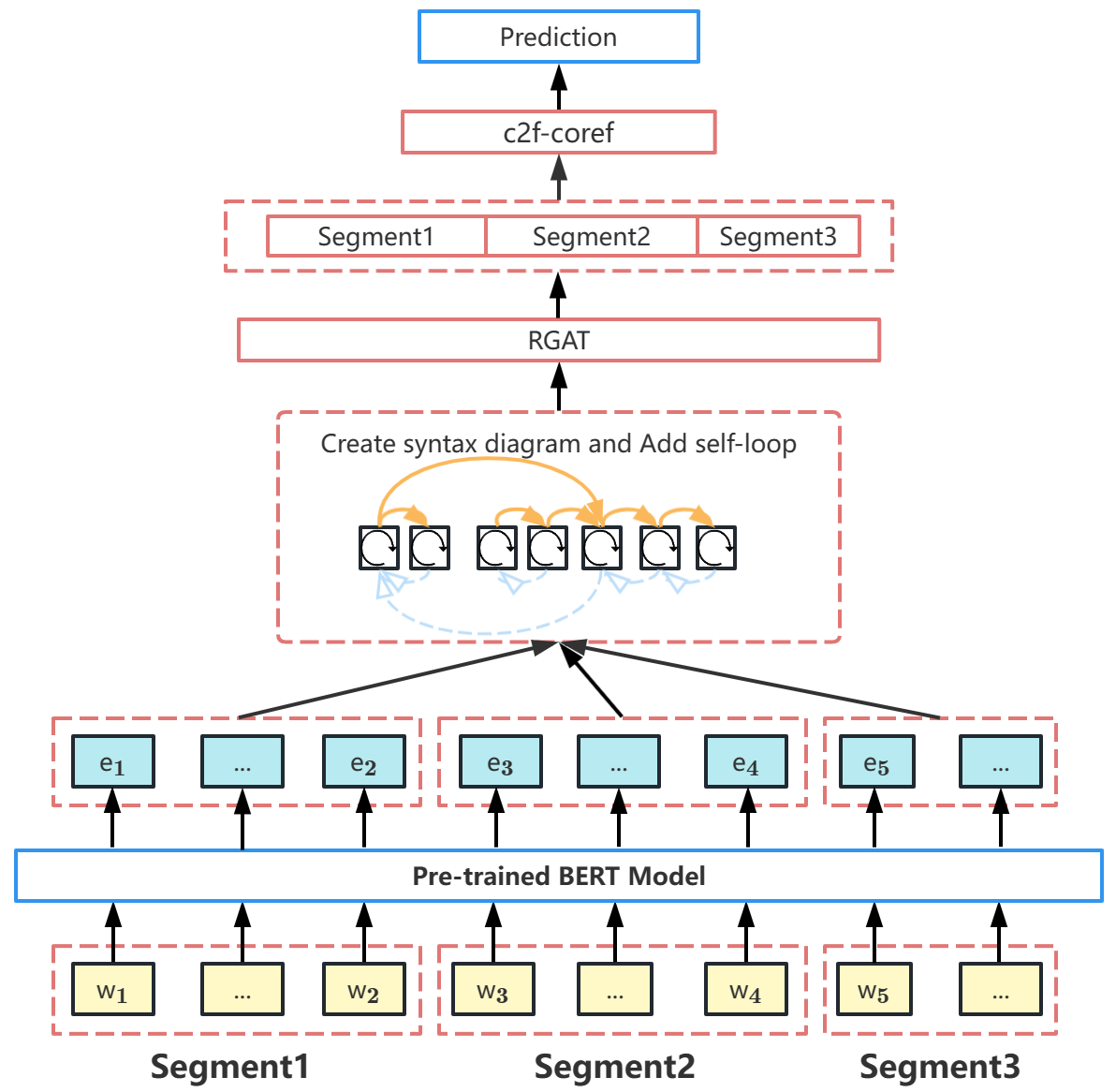}}
\caption{RGAT-with-BERT + c2f-coref model for OntoNotes 5.0 dataset.}
\label{fig5}
\end{figure}

\noindent \textbf{Implementation Detail.} We also evaluate our models on the OntoNotes 5.0 dataset \cite{b25}. Different from the GAP dataset, the OntoNotes 5.0 examples are considerably longer and typically require multiple segments to read the entire document. We replace the entire BERT-large transformer (embeddings as input) in c2f-coref with the RGAT-with-BERT + c2f-coref (our model). The general model architecture is shown in Fig.~\ref{fig5}.

We treat the first and last word-pieces (concatenated with the attended version of all word pieces in the span) as span representations. As shown in Fig.~\ref{fig5}, each sample is split into three segments in our model, and each segment will be used as input to the BERT model. After passing through the RGAT model, the output will be passed through the c2f-coref architecture to get the predicted results. 

\noindent \textbf{Results.} We compare the RGAT-with-BERT + c2f-coref system with three main baselines: (1) the original ELMo-based c2f-coref system \cite{b26}, and (2) its predecessor, e2e-coref \cite{b22} and (3) the BERT-based c2f-coref system \cite{b27}. Experimental results show that our work achieves a large improvement over baseline models, which is shown in TABLE~\ref{tab3}. (P means precision, R means recall, F1 means F1-score).

\begin{table*}[htbp]
\caption{Results on OntoNotes5.0 Dataset}
\begin{center}
\begin{tabular}{|l|lll|lll|lll|l|}
\hline
 & \multicolumn{3}{c|}{$MUC$}                      & \multicolumn{3}{c|}{$B^3$}                        & \multicolumn{3}{c|}{$CEAF_{\varphi 4}$}                        &               \\
                  & P             & R             & F1             & P             & R             & F1             & P             & R             & F1             & \textbf{Avg.F1}             \\ \hline
\textbf{e2e-coref \cite{b22}}         & 78.4          & 73.4          & 75.8          & 68.6          & 61.8          & 65.0          & 62.7          & 59.0            & 60.8          & 67.2          \\
\textbf{c2f-coref \cite{b26}}         & 81.4          & 79.5          & 80.4          & 72.2          & 69.5          & 70.8          & 68.2          & 67.1          & 67.6          & 73.0          \\
\textbf{Fei et al. \cite{b28}}                & 85.4          & 77.9          & 81.4          & \textbf{77.9} & 66.4          & 71.7          & 70.6          & 66.3          & 68.4          & 73.8          \\
\textbf{EE \cite{b29} }              & 82.6          & 84.1          & 83.4          & 73.3          & \textbf{76.2} & 74.7          & 72.4          & \textbf{71.1} & 71.8          & 76.6          \\ \hline
\textbf{BERT-base + c2f-coref \cite{b27}}       & 80.2          & 82.4          & 81.3          & 69.6          & 73.8          & 71.6          & 69.0          & 68.6          & 68.8          & 73.9          \\
\textbf{BERT-large + c2f-coref \cite{b27}}       & 84.7          & 82.4          & 83.5          & 76.5          & 74.0          & 75.3          & 74.1          & 69.8          & 71.9          & 76.9          \\ \hline
\textbf{RGAT-with-BERT + c2f-coref(ours)}   & \textbf{85.7} & \textbf{82.5} & \textbf{84.3} & 77.6          & 73.9          & \textbf{76.5} & \textbf{75.2} & 70.0          & \textbf{72.8} & \textbf{77.7} \\ \hline
\end{tabular}
\label{tab3}
\end{center}
\end{table*}

TABLE~\ref{tab3} shows that RGAT-with-BERT + c2f-coref outperforms the BERT-large + c2f-coref model on English by 0.8\% on the OntoNotes 5.0 Dataset. The main evaluation metric is the average F1 of three metrics -- $MUC$, $B^3$ and $CEAF_{\varphi 4}$ on the test set. Given how gains on coreference resolution have been hard to come by as evidenced by baseline models in TABLE~\ref{tab3}, our model is still a considerable improvement. It is noted that compared with BERT, we only add a relatively small number of parameters, which can get a more obvious effect on the reference resolution task. Due to the limitation of computing resources, we did not tune high parameters further. In view of the experimental results, we believe that the syntactic structure can indeed help the model to further understand the coreference resolution task.

\section{Conclusion and Discussion}
\subsection{Conclusions}
The experiment results show that with the help of sentence syntactic dependency information, using the output representations of BERT pre-trained model, RGAT can further learn embedding representations that are more conducive to the task of pronoun resolution and improve the performance of this task.

 “Gender Bias in Natural Language Processing (GeBNLP) 2019 Shared Task” is a competition to build a common reference parsing system on the GAP dataset. We employ a combination of BERT and our proposed graph neural network RGAT model to participate in this task. The RGAT model is used to digest the syntactic dependency graph, and further extract syntactic-related information from the output features by BERT, which helps us improve the accuracy of this task. Our model significantly improves the F1-score of the task from the previous best of 80.3\% to 82.5\% (an improvement of 2.2\%). Compared with BERT plus a fully connected layer, the accuracy of fine-tuning the fully connected layer is only 78.5\%, and our model has a 4.0\% improvement. The results show that without fine-tuning the BERT model, the syntactic dependency graph can significantly improve the performance of the referencing problem.

 For another classic dataset -- OntoNotes 5.0, the comparison results of RGAT-with-BERT +c2f-coref VS. baseline BERT-large + c2f-coref indicates that syntactic structure might better encode longer contexts. These observations suggest that future research in pretraining methods may look at more effectively encoding document-level context including syntactic structure. Modelling pronouns, especially in the context of dialogue, still has a lot of potential for our model. Although our advantages are not very obvious, partly because we are limited to the c2f-coref architecture, we believe syntactic structure can effectively help our model achieve more comparable results for document modelling if we can design a suitable architecture for our model. Lastly, through the overview of training samples and our model outputs, a considerable number of errors suggest that our model is also still unable to resolve cases requiring mention paraphrasing like \cite{b27}, especially considering that such learning signal is rather sparse in the training set. However, a few of these errors have been reduced. We think our model has the possibility to solve this problem.

\subsection{Discussion}
In fact, our work provides an alternative paradigm for such similar coreference tasks or for those tasks that need to mine the role of syntactic information. Our work demonstrates it is not so necessary to fine-tune the entire BERT model and save a unique BERT model parameter for each task. Our proposed paradigm simply changes the classification header of the BERT model to graph neural networks with syntactic dependencies, and then fine-tunes the new classification header to obtain better results.

Our work is to solve the problems encountered by the current large pre-trained language model from a new perspective, that is, the pre-trained language model is too large, and it is necessary to save a new fine-tuned model parameter for each downstream task. For example, models such as LoRA \cite{b8}, AdapterBias \cite{b30}, and LLM-Adapters \cite{b31} all reduce the amount of parameter required for fine-tuning the model on the model itself and the output of each layer. And other models that incorporate traditional machine learning methods \cite{b4} \cite{b13} do not achieve competitive results. By comparison, our work is to use the output features of the pre-trained language model, but not to change the parameters of the model itself and the output features. Our work shows that changing the classification head can effectively reduce the amount of parameter for fine-tuning the pre-trained model and greatly improve the recognition accuracy of the task.

\subsection{Limitation and Future Direction}
Our models and experiments have shown that syntactic dependency information plays a significant role in reference resolution tasks and that syntactic structure can optimize the embedded representation of large language models. However, there are three problems to solve in the future. (1) There is no evidence of the role of syntactic structures for other NLP tasks. (2) In this paper, supervised learning is used to optimize the embedded representations of BERT. It is a future direction to explore the unsupervised representation learning that combines BERT with syntactic structure. (3) When training, we can first save the features to the hard disk, but in inference, we do not store embedded features, so how to optimize the inference time is also a future problem.

\section*{Acknowledgment}
This research was funded by Shanghai Philosophy and Social Sciences Planning Project, grant number 2020BGL009.

\end{document}